\documentclass{article}
\usepackage{arxiv}

\usepackage[utf8]{inputenc} 
\usepackage[T1]{fontenc}    
\usepackage{hyperref}       
\usepackage{url}            
\usepackage{booktabs}       
\usepackage{amsfonts}       
\usepackage{nicefrac}       
\usepackage{microtype}      

\usepackage{graphicx}
\usepackage[round]{natbib}
\usepackage{esvect}
\usepackage{amsmath}
\usepackage[tight,footnotesize]{subfigure}
\usepackage{appendix}
\usepackage{lscape}
\usepackage{textcomp}
\usepackage{amssymb}
\usepackage{lscape}
\usepackage[noend]{algpseudocode}
\usepackage[boxruled]{algorithm2e}
\usepackage{hhline}

\title{A self-organizing fuzzy neural network for sequence learning}

\author{
  Armin Salimi-Badr \\
  Faculty of Computer Science and Engineering\\
  Shahid Beheshti University\\
  Tehran, Iran \\
  \texttt{a\_salimibadr@sbu.ac.ir} \\
   \And
 Mohammad Mehdi Ebadzadeh \\
  Department of Computer Engineering\\
  Amirkabir University of Technology\\
  Tehran, Iran \\
  \texttt{ebadzadeh@aut.ac.ir} \\
}
\begin{document}
\maketitle
\begin{abstract}
In this paper, a new self-organizing fuzzy neural network model is presented which is able to learn and reproduce different sequences accurately. Sequence learning is important in performing skillful tasks, such as writing and playing piano. The structure of the proposed network is composed of two parts: 1- \textit{sequence identifier} which computes a novel sequence identity value based on initial samples of a sequence, and detects the sequence identity based on proper fuzzy rules, and 2- \textit{sequence locator}, which locates the input sample in the sequence. Therefore, by integrating outputs of these two parts in fuzzy rules, the network is able to produce the proper output based on current state of the sequence. To learn the proposed structure, a gradual learning procedure is proposed. First, learning is performed by adding new fuzzy rules, based on coverage measure, using available correct data. Next, the initialized parameters are fine-tuned, by gradient descent algorithm, based on fed back approximated network output as the next input. The proposed method has a dynamic structure which is able to learn new sequences online. The proposed method is used to learn and reproduce different sequences simultaneously which is the novelty of this method. \\
\textbf{Keywords:} Self-Organizing Fuzzy Neural Networks . Sequence Learning . Gradual Learning
\underline{The full version of this preprint is accepted for publication in IEEE Transactions on Cybernetics as}\\ \underline{"A Novel Self-Organizing Fuzzy Neural Network to Learn and Mimic Habitual Sequential Tasks," with}\\  $DOI:10.1109/TCYB.2020.2984646$
\end{abstract}


\section{Introduction}
\label{section1}
Fuzzy neural network (FNN) \citep{ANFIS,Kasabov01,GDFNN,SOFMLS,Ebadzadeh09,Ebadzadeh15,Ebadzadeh2017,Rubio15,Ma16,Ganji2017,Han2017modeling,Han2018self} is a hybrid method that combines the learning capability of a neural network with the interpretability of a rule-based fuzzy system. It is proved that FNNs are universal approximators \citep{Kosko94,Ying98,Zeng00,Ebadzadeh15,Das15}. The functionality of FNN as a powerful tool is investigated in different applications including data mining, signal processing, system modeling, fault diagnosis, robotics and control \citep{Ebadzadeh08,Han14,Kim15,AsadiEydivand2015,Jahromi2016,Salimi-Badr2017,Qiao2018adaptive,Han2018self,Pang2018,Khodabandelou2019,Xu2019fuzzy,Wen2019infrared}.

Two important issues in constructing an FNN are the \textit{structure identification} and \textit{parameter estimation} \citep{SOFMLS}. The structure identification is related to determining the number of fuzzy rules and the parameter estimation concerns deriving proper values of parameters to construct a reliable system \citep{Ebadzadeh09}.

In \citep{ANFIS}, the well-known adaptive network-based fuzzy inference system (ANFIS) is proposed. The gradient descent method is used to learn the network parameters. For structure identification, each input variable space is partitioned uniformly into fuzzy sets. In methods such as ANFIS, it is assumed that all data are available at once.

Some previous methods, considering the dynamic nature of real-world problems, utilize online structure identification \citep{DFNN,DENFIS,FAOSPFNN,SOFMLS,GPFNN,Han14,Malek11,Salimi-Badr2017,Han2018self}. In \citep{DFNN}, a fuzzy neural network has been proposed with a dynamic structure. In order to perform \textit{structure identification}, a hierarchical online self-organizing learning mechanism is used. Generally, such methods try to avoid producing extra rules by selecting them based on some appropriate criteria \citep{FAOSPFNN,Pizzileo12}.

A real-world application with dynamic nature is sequence learning. Vertebrates, including human-beings, can learn and reproduce different sequences of actions, and generate different patterns subsequently. This ability is necessary to learn and perform skillful movements, such as playing piano, writing, or singing.

Although previous dynamic FNNs, such as \citep{GDFNN,DFNN,DENFIS,FAOSPFNN,SOFMLS,Malek11,AsadiEydivand2015,Salimi-Badr2017}, are designed to work in dynamic environments, such as learning identity functions of a dynamic system, they are not designed to learn and reproduce \textit{different} sequences. Sequence learning for ordinary dynamic FNNs is even more complex, when different sequences have similar parts.

In this paper, a new Self-organizing FNN architecture is proposed for sequence learning. The proposed structure composed of two parts, \textit{sequence identifier} and \textit{sequence locator}. The first part receives only the initial samples of a sequence, to identify the sequence pattern among different learned sequences. The second part is designed to locate the current state in the current sequence. By combining the outputs of these parts and utilizing them in fuzzy rules, the network is able to identify the sequence and the current location, to produce the next sample properly.

The proposed network is utilized to learn two different sequences with a similar part. Since the network is designed to identify each sequence based on initial samples by \textit{sequence identifier} part, it can learn these trajectories precisely. It is also used to learn different patterns and generate them periodically, as a pattern generator. As a real world problem, different hand writing character trajectories are learned and reproduced by the proposed network successfully.

The rest of this paper is organized as follows. In Section \ref{section2}, the structure of the proposed architecture, initialization method and fine-tuning algorithm are discussed in details. Section \ref{section3} reports the effectiveness of the proposed method in learning different sequences. Finally, conclusions are presented in Section \ref{section4}.

\section{Proposed Method}
\label{section2}
In this section, the main ideas underlying the proposed algorithm are presented. After describing the proposed structure of the Fuzzy Neural Network (FNN), the initialization algorithm and the fine-tuning method are presented.

\subsection{Proposed Architecture}

\begin{figure}
\centering
\includegraphics[width=5in]{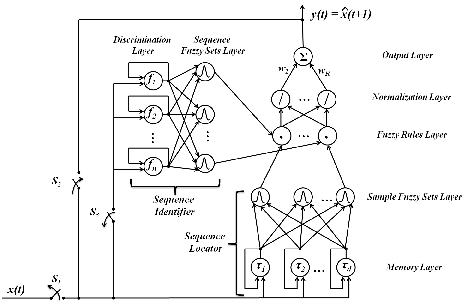}
\caption{The proposed architecture for sequence learning. The network is composed of two parts, \textit{sequence identifier} and \textit{sequence locator}. The \textit{discrimination layer} contains recurrent neurons which calculate a discriminative identity number for each sequence by using $f_i$s which are polynomial integrating functions. \textit{Memory layer} contains recurrent neurons which play the role of low-pass filters with different time-constants $\tau_j$. \textit{Sequence fuzzy sets layer} and \textit{Sample fuzzy sets layer} are composed of RBF neurons with multivariate Gaussian functions. The \textit{sequence identifier} identifies the sequence and the \textit{sequence locator} detect the location of the current sample. Next, their outputs are combined in the \textit{fuzzy rules layer}. After applying normalization in the \textit{normalization layer}, next sample of the sequence is approximated ($\hat{x}(t+1)$) as the network output ($y(t)$). In the output layer, $w_1$ to $w_R$ are output weights for each fuzzy rule. It is assumed that the network would identify and produce the sequence after $T$ samples. Therefore, there are three switches $S_1$ to $S_3$. Initially, $S_1$ and $S_3$ are closed and the correct sequence sample $x(t)$ is received by the network. After $T$ samples, these switches become open and $S_2$ would be closed. Thus, the \textit{sequence locator} part receives the previous network output as the input, and the \textit{sequence identifier} part receives no more inputs. Indeed, the output of the this part would remain constant after $T$ samples.}
\label{fig_1}
\end{figure}

The structure of the proposed FNN is shown in Fig. \ref{fig_1}. The network is composed of two parts: 1- \textit{sequence identifier}, and 2- \textit{sequence locator}. Each part contains two distinct layers. The \textit{sequence identifier} receives $T$ initial samples of a sequence, and identifies the sequence. The second part, \textit{sequence locator}, receives the current sample of a sequence ($x(t)$) and detects its location in the sequence. The outputs of these two parts are integrated into a fuzzy rule firing strength, which is  utilized to calculate the next sample of the identified sequence ($\hat{x}(t+1)$). It is assumed that the network would identify the sequence and produce the samples of sequence after $T$ samples. Therefore, there are three switches, $S_1$ to $S_3$ in the proposed circuit to manage the information flow in the network. Initially, $S_1$ and $S_3$ are closed and the network receives the correct sequence sample $x(t)$. After receiving $T$ samples, these switches become open and $S_2$ would become closed. Thus, the \textit{sequence locator} part receives the previous network output as the input, and the \textit{sequence identifier} part does not receive any inputs. Indeed, the output of the this part would remain constant after $T$ samples.

Totally, the proposed network is composed of seven distinct layers. These layers are the \textit{discrimination layer}, the \textit{sequence fuzzy sets layer}, the \textit{memory layer}, the \textit{sample fuzzy sets layer}, the \textit{fuzzy rules layer}, the \textit{normalization layer} and finally the \textit{output layer}. The output layer plays a role in determining the consequent parts of the rules, and the others serve as premise parts. We denote $l$ as the layer number, and $o^{(l)}_i$
as the output of the $i^{th}$ node of the $l^{th}$ layer.

The details of each layer are as follows:
\begin{enumerate}
\item \textit{The discrimination layer}: The function of this layer is to calculate discriminative identity number for each sequence. This layer receive $T$ samples of the sequence. It contains neurons with polynomial function $f_i$. If $x(t)$ is the input in $t^{th}$ time instant, the function of $i^{th}$ neuron in this layer is as follows:
\begin{equation}
    o^{(1)}_i(t) = \left\{\begin{array}{ll}
    o^{(1)}_i(t-1) + x^i(t); & t \leq T \\
    o^{(1)}_i(t-1); & t > T
    \end{array}\right.
\label{eq_1}
\end{equation}
Therefore, after $T$ samples, the output of its $i^{th}$ neuron would be $\sum_{t=1}^Tx^i(t)$.

\item \textit{The sequence fuzzy sets layer}: In this layer, multivariate fuzzy sets are defined for different identity numbers which are calculated in the previous layer. Indeed, each neuron of this layer receives a vector which contains all outputs of the previous layer ($o^{(1)}$). The output of the $i^{th}$ neuron of this layer is calculated as follows:
\begin{equation}
    o^{(2)}_i = e^{-(o^{(1)}-c1_i)^T\Sigma_1^{-1}(o^{(1)}-c1_i)}
\label{eq_2}
\end{equation}
where, here $c1_i$ is the vector of $i^{th}$ fuzzy set center and $\Sigma_1$ is a diagonal matrix contains width of fuzzy set for each input dimension as follows:
\begin{equation}
\Sigma_1 = \left[\begin{array}{cccc}
        \sigma_1^2 & 0 &\cdots & 0 \\
        0 & \sigma_1^2 & \cdots & 0 \\
        \vdots & \vdots & \ddots & \vdots \\
        0 & 0 &\cdots & \sigma_1^2 \end{array}\right]
\label{eq_3}
\end{equation}

\item \textit{The memory layer}: Each neuron of this layer is a discrete low-pass filter with a different time constant $\tau_i$. Indeed, this layer output is a vector which contains current and $d$ previous samples of the sequence and its $i^{th}$ neuron represents the $(t-i)^{th}$ sample. The output of the $i^{th}$ neuron of this layer is calculated as follows:
\begin{equation}
    o^{(3)}_i(t) = \left\{\begin{array}{ll}
    \lambda_i o^{(3)}_i(t-1) + (1-\lambda_i)x(t); & t \leq T \\
    \lambda_i o^{(3)}_i(t-1) + (1-\lambda_i)y(t-1); & t > T
    \end{array}\right.
\label{eq_4}
\end{equation}
where $\lambda_i = \frac{i}{i+1}$. It is assumed that the number of neurons in this layer, ($d$), is lower than number of initial samples $T$.

\item \textit{The sample fuzzy sets layer}: The function of this layer is similar to the second layer, but its input is the output of the third layer. Therefore, the output of its $i^{th}$ neuron is calculated as follows:
\begin{equation}
    o^{(4)}_i = e^{-(o^{(3)}-c2_j)^T\Sigma_2^{-1}(o^{(3)}-c2_j)}
\label{eq_5}
\end{equation}
where, here $c2_i$ is the vector of $i^{th}$ fuzzy set center and $\Sigma_2$ is a diagonal matrix similar to $\Sigma_1$.

\item \textit{The fuzzy rules layer}: Each neuron of this layer represents a fuzzy rule. Its neurons compute a membership value based on applying the \textit{T-Norm} operator on previous layer's outputs. By considering the inner product as the \textit{T-Norm} operator, the membership function of $i^{th}$ fuzzy rule is defined as shown bellow:
\begin{equation}
    o^{(5)}_i = \mu_i =  o^{(2)}_j.o^{(4)}_k
\label{eq_6}
\end{equation}
where $\mu_i$ is the membership value of the current sequence and its current location to the $i^{th}$ fuzzy rule.

\item \textit{The normalization layer}: Calculated membership values are normalized in this layer. The output of $i^{th}$ neuron of this layer is calculated as follows:
\begin{equation}
    o^{(6)}_i = \phi_i =  \frac{o^{(5)}_i}{\sum_{j=1}^R o^{(5)}_j}
\label{eq_7}
\end{equation}
where $\phi_i$ is the normalized membership value of the current sequence and its current location to the $i^{th}$ fuzzy rule, and $R$ is the number of fuzzy rules.

\item \textit{The output layer}: There is one linear neuron for each output, in the last layer that provides the final output of the network. The output of this neuron is the inner product of the input vector with a weight vector $(W)$, which plays the role of the consequent part parameters. The final output of the network is computed as follows:
\begin{equation}
o^{(7)} = y(t) = \hat{x}(t+1) = W^T.o^{(6)}
\label{eq_8}
\end{equation}

\end{enumerate}

 The advantage of the proposed architecture is to have two distinct parts for sequence identification and location. Indeed, this architecture would be able to recognize different sequences (for example different characters in handwriting) by the \textit{sequence identifier} part, and also the current state in each sequence by the \textit{sequence locator}. Therefore, it is proper for sequence learning.

\subsection{Initialization}
Considering the dynamic nature of a sequence, a dynamic learning method is proposed for structure learning. During this phase of learning, switch $S_1$ is closed and $S_2$ is open. Switch $S_3$ is closed for first $T$ samples and after receiving $T$ samples, this switch would be open. Therefore, during this phase of learning, the architecture utilize correct sequence samples ($x(t)$).

After receiving $T$ samples, the learning procedure starts with considering the current output of the \textit{memory layer} as the center of the first fuzzy set in the \textit{sample fuzzy sets layer}, and the output of the \textit{discrimination layer} as the center of the first fuzzy set in the \textit{sequence fuzzy sets layer}. To form the first fuzzy rule, these two fuzzy sets are joint in the \textit{fuzzy rules layer}. To initialize the output weight for the first fuzzy rule, the desired output of the current sample (x(t+1)) is considered as the output weight.

\begin{figure}
\centering
\begin{tabular}{c}
    \subfigure[][]{\includegraphics[width = 3.1in]{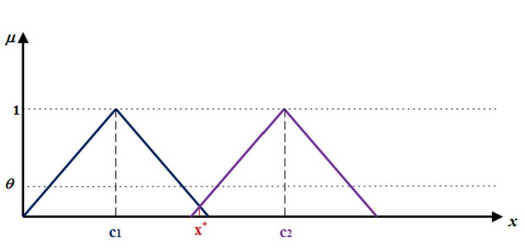}}\\
    \subfigure[][]{\includegraphics[width = 3.1in]{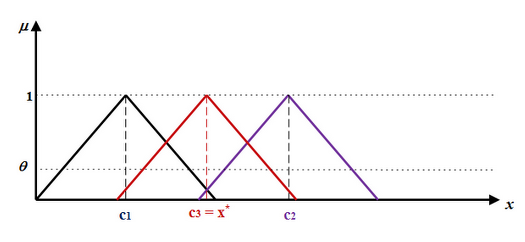}}
\end{tabular}
\caption{A simple example of adding a new fuzzy set in a one-dimensional input space. \textbf{a} The input signal is analyzed by two fuzzy sets with centers $c_1$ and $c_2$. The total coverage for the sample $x^*$ is lower than a predefined threshold $\theta$. \textbf{b} A new fuzzy set, which its center is $x^*$, has been added.}
\label{fig_2}
\end{figure}

Next, all samples of the first sequence are fed to the network. If the current set of fuzzy sets in \textit{sample fuzzy sets layer} do not cover each sample $x(t)$, a new fuzzy set is added to this layer that its center is $x(t)$. Indeed, if the sum of fuzzy sets firing strength ($\sum_{i=1}^{m}o4_i$) is lower than a predefined threshold ($\theta_1$), a fuzzy set is added to the \textit{sample fuzzy sets layer}. Fig. \ref{fig_2} presents a schematic example of this concept. For the first sequence, each new fuzzy set in the \textit{sample fuzzy sets layer} would be joint with the fuzzy set in the \textit{sequence fuzzy sets layer} to from fuzzy rules. To initialize output weight, the desired value of the next sample ($x(t+1)$) is considered as the output weight of the new fuzzy rule.

After learning the first sequence, the other sequences are fed to the network. If the current set of fuzzy sets in \textit{sequence fuzzy sets layer} could not cover the new identity number for each sequence, a new fuzzy set would be added to this layer which its center would be the current identity number. If a new fuzzy set is added to this layer, it is necessary to form new fuzzy rules for the new sequence. Now, if the current set of fuzzy sets in \textit{sample fuzzy sets layer} cover sample $x(t)$, the fuzzy set with highest firing strength ($argmax(o^{(4)}_i)$) would be joint with the added fuzzy set in \textit{sequence fuzzy sets layer} to form a new fuzzy rule. Else, a new fuzzy set is added to \textit{sample fuzzy sets layer} that its center is $x(t)$. Then, to form new fuzzy rule, this new fuzzy set is joint with the new fuzzy set in \textit{sequence fuzzy sets layer}. This algorithm is summarized in Algorithm \ref{alg_1}.

\begin{algorithm}[tbp]\footnotesize
\caption{Initialization}\label{alg_1}
$seq \gets 1$;

$m \gets 0$; \tcc{Number of fuzzy sets in sample FSL}

$p \gets 0$; \tcc{Number of fuzzy sets in sequence FSL}

$R \gets 0$; \tcc{Number of rules}

$rules \gets []$; \tcc{Joint fuzzy sets forming rules}

\While{seq $\leq$ seqNum}{

    $newSeq \gets true$;

    \For{$t \gets 1 \quad \KwTo \quad T$}{
        \text{Update $o^{(1)}$ (eq. (\ref{eq_1}))}
        }
    $coverage_1 = \sum_{i=1}^po^{(2)}_i$;

    \If{newSeq $\&\&$ $coverage_1$ $\leq$ $\theta_1$}{
    \tcc{Add new fuzzy set in  sequence FSL:}

    $p \gets p + 1$;

    $c1_p \gets o1$;

    $newSeq \gets false$;

    }

    \For{$t \gets T+1 \quad \KwTo \quad T_{f}$}{
        \text{Update $o^{(3)}$ based on $x(t)$ (eq. (\ref{eq_4}))}

        \text{Update $o^{(4)}$ based on x(t) (eq. (\ref{eq_5}))}

         $coverage_2 \gets \sum_{i=1}^mo^{(5)}_i$;

        \If{$coverage_2$ $\leq$ $\theta_2$}{
        \tcc{Add new fuzzy set in  sample FSL:}

        $m \gets m + 1$;

        $c2_m \gets o^{(5)}$;

        \tcc{Add new rule:}

        $R \gets R+1$;

        $rules[R,1] \gets p$;

        $rules[R,2] \gets m$;

        $w_R \gets x(t+1)$;

        }

         \tcc{Check the coverage of rules:}

        $M \gets argmax(o^{(4)})$;

        $ruleRequired \gets true$;

        \For{$i \gets 1 \quad \KwTo \quad R$}{
        \If{rules[i] = [p,M]}{

        $ruleRequired \gets false$;

        }
        }

        \If{ruleRequired}{

         \tcc{New rule should be added:}

                $R \gets R+1$;

                $rules[R,1] \gets p$;

                $rules[R,2] \gets M$;

                $w_R \gets x(t+1)$;

        }

        }

        $seq \gets seq + 1$;
}
\end{algorithm}

\subsection{Fine-tuning}
After initialization, the circuit would work in normal operational mode: For $t \leq T$, switches $S_1$ and $S_3$ are closed and $S_2$ is open; for $t > T$, $S_2$ is closed and $S_1$ and $S_2$ are open. Therefore, the approximated value is fed back to the model as the next input after $T$ samples.

In this phase of learning, only output weights are fine-tuned based on the network output, using gradient descent algorithm as follows:
\begin{equation}
\begin{array}{ll}
    w_i &= w_i - \eta\frac{\partial E(t)}{\partial w_i} \\
    &= w_i - \eta\frac{\partial (y(t)-x(t+1))^2}{\partial w_i}\\
    &= w_i - \eta(y(t)-x(t+1))\frac{\partial y(t)}{\partial w_i}\\
    &= w_i - \eta(y(t)-x(t+1))\phi_i
\end{array}
    \label{eq_9}
\end{equation}
where $E(t)$ is the squared error in time instant $t$, and $\eta$ is the learning step value.

To ensure that the propagated error due to recurrent nature of network is controlled, a gradual fine-tuning procedure is applied. This gradual fine-tuning method utilizes eq. (\ref{eq_9}) sequentially for each sample. After reducing the error for sample $x(t)$, iteratively, to lower than a predefined threshold $\theta_3$, the algorithm is used for the next sample $x(t+1)$. This algorithm is summarized in Algorithm \ref{alg_2}.

\begin{algorithm}[tbp]\footnotesize
\caption{Fine-tuning Algorithm}\label{alg_2}

\For{$seq \gets 1 \quad \KwTo \quad seqNum$}{

    $\eta \gets \eta_0$;

    \For{$t \gets 1 \quad \KwTo \quad T$}{

        \text{Update o1 to o7 based on x(t) (eqs. (\ref{eq_1}) to (\ref{eq_8}))}

        }

\For{$iter_1 \gets 1 \quad \KwTo \quad iter_{max}$}{

    \For{$t \gets T+1 \quad \KwTo \quad T_{f}$}{

        $iter_2 \gets 1$;

        $E \gets (y(t)-x(t+1))^2$;

        \While{$iter_2$ $\leq$ $iter_{max}$ $||$ $E \leq \theta_3$}{

             \For{$i \gets 1 \quad \KwTo \quad R$}{

             \text{Update $w_i$ (eq. (\ref{eq_9}))}

            }

            \text{Update o1 to o7} \text{\quad based on y(t) (eqs. (\ref{eq_1}) to (\ref{eq_8}))}

            $E \gets (y(t)-x(t+1))^2$;

            \tcc{$\beta \in (0,1)$ is a discounting factor}

            $\eta \gets \beta.\eta$;

        }

    }
}
}
\end{algorithm}

\section{Experiments}
\label{section3}
In this section, the performance of the proposed network in sequence learning is investigated. First, it is used to learn two intersected sequences and its performance is compared to the performance of an ordinary dynamic FNN (\citep{Salimi-Badr2017}). Next, the proposed network is utilized as a pattern generator to learn and generate four types of periodic signals. Finally, this network is utilized to learn different character trajectories extracted from UCI character trajectories data set\footnote{https://archive.ics.uci.edu/ml/datasets/Character+Trajectories} \citep{williams2006,williams2007,williams2008}. This data set includes multiple, labelled samples of pen tip trajectories recorded whilst writing individual characters. All samples are from the same writer. Its sample have three dimensions including $x$, $y$, and pen tip force. In this paper, only two dimensions $x$ and $y$ are used. To show the ability of proposed FNN to encounter noise and uncertainty, random noise is added to the input samples.

\subsection{Intersected sequences}
\begin{figure}[!t]
    \centering
    \includegraphics[width=2in]{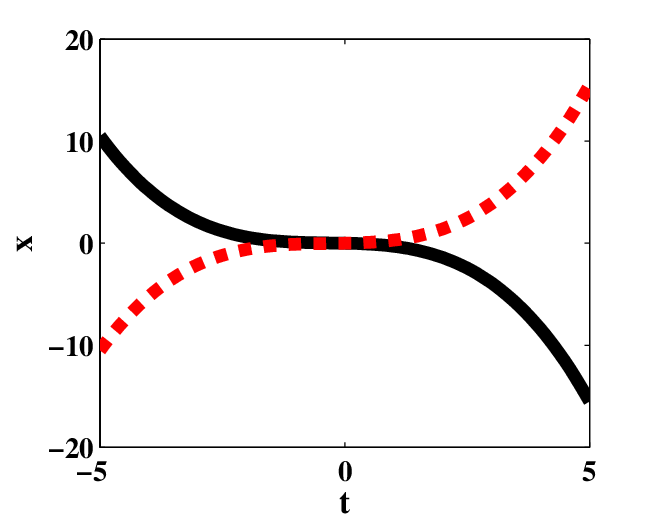}
    \caption{Two intersected sequences with a similar part.}
    \label{fig_3}
\end{figure}

In this experiment, the ability of the proposed network is compared to an ordinary self-organizing fuzzy neural network structure (\citep{GDFNN,Malek11,Salimi-Badr2017}) to learn two different intersected sequences. Indeed, these two sequences are begun from different points, and after some samples, they  converge to a similar curve, and then, they diverge in different directions (Fig. \ref{fig_3}).

\begin{table}[t!]
    \caption[c]{Parameters values for learning two intersected functions}
    \begin{center}
    \begin{tabular}{|c|c|c|c|c|c|c|c|}
        \hline
         $iter_{max}$ & $\theta_1$ & $\theta_2$ & $\theta_3$ & $\sigma$ & $T$ & $d$ &$n$\\
         \hline
         20 & 0.3 & 0.3 & 0.01 & 0.1 & 10 & 5 & 1  \\
         \hline
    \end{tabular}
    \end{center}
 \label{table_1}
 \end{table}

\begin{figure*}
\centering

\begin{tabular}{cc}
    \subfigure[][]{\includegraphics[width = 1.5in]{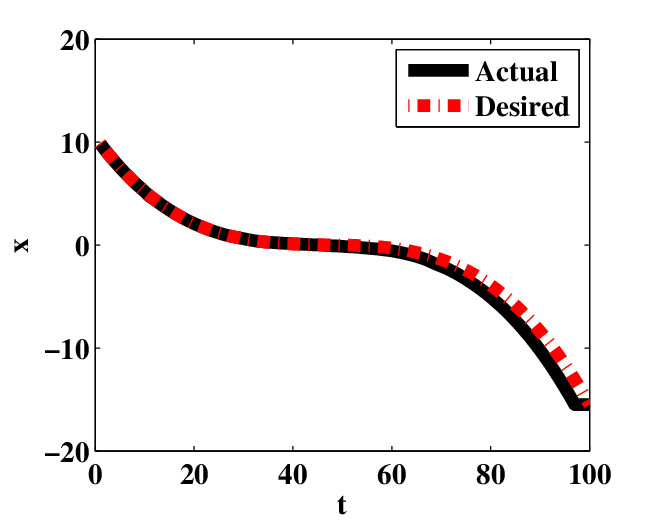}} & \subfigure[][]{\includegraphics[width = 1.5in]{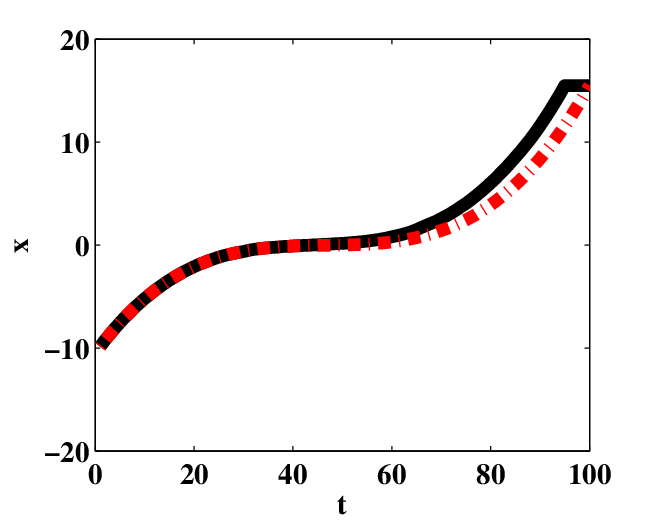}}  \\
    \subfigure[][]{\includegraphics[width = 1.5in]{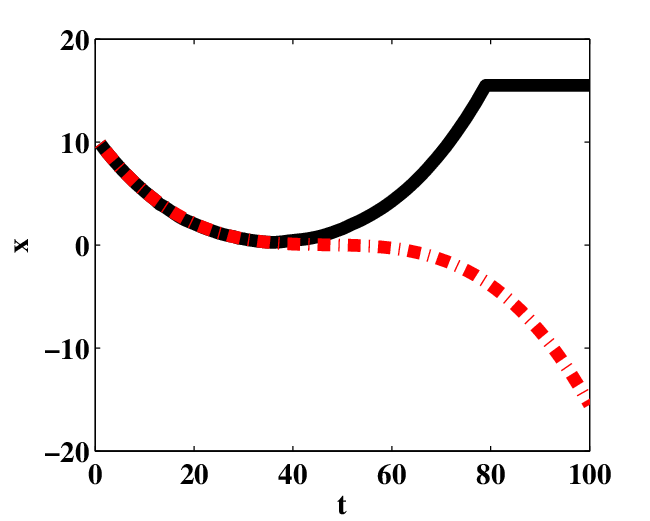}} & \subfigure[][]{\includegraphics[width = 1.5in]{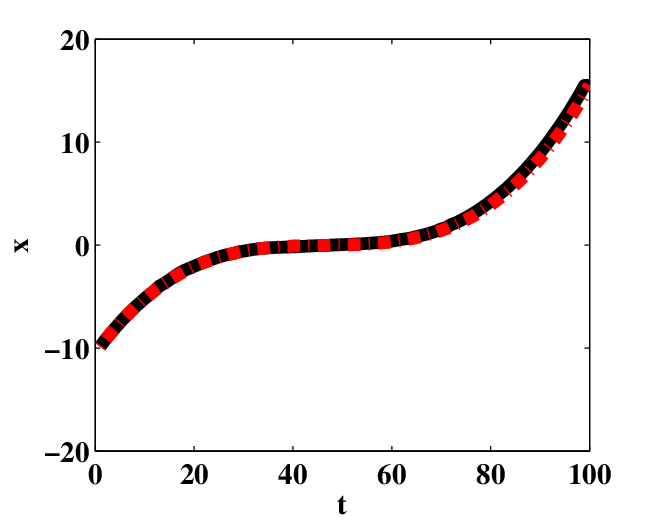}}
\end{tabular}
\caption{Comparison between the proposed architecture and the ordinary one in learning two trajectories with intersected region. \textbf{a} Proposed network output for the first sequence. \textbf{b} Proposed network output for the second sequence. \textbf{c} The output of an ordinary self-organizing network architecture (GDFNN) for the first sequence. \textbf{d} The output of an ordinary self-organizing network architecture (GDFNN) for the second sequence.}
\label{fig_4}
\end{figure*}

Table \ref{table_1} shows different parameters values for both approaches and Fig. \ref{fig_4} compares outputs of the proposed network and the ordinary one. According to Fig. \ref{fig_4}, the proposed network is able to learn and generate both sequences, but the ordinary network could learn just one sequence, due to the intersected region. Therefore, the proposed network is able to learn and generate such sequences by utilizing a small window of the previous samples.

\subsection{Pattern generation}
One application of the proposed network is to be utilized as a pattern generator. In this experiment, the proposed network is used to learn and generate periodic patterns. Four patterns, including sine, square, triangle, and sawtooth waves are used as target sequences. The network should learn to produce one period of all patterns based on receiving the samples of the first period. Next, it would be able to reproduce them periodically, by receiving just the first period. Table \ref{table_2} shows the values of network parameters. After learning, the number of  fuzzy sets of the \textit{sequence fuzzy sets layer} is 4 and number of fuzzy sets of the \textit{sample fuzzy sets layer} is 57.

\begin{table}[t]
    \caption[c]{Parameters values for learning different patterns}
    \begin{center}
    \begin{tabular}{|c|c|c|c|c|c|c|c|}
        \hline
         $iter_{max}$ & $\theta_1$ & $\theta_2$ & $\theta_3$ & $\sigma$ & $T$ & $d$ &$n$\\
         \hline
         20 & 0.4 & 0.1 & 0.01 & 0.1 & 20 & 20 & 2  \\
         \hline
    \end{tabular}
    \end{center}
 \label{table_2}
 \end{table}

\begin{figure}
\centering
\begin{tabular}{cc}
    \subfigure[][]{\includegraphics[width = 1.5in]{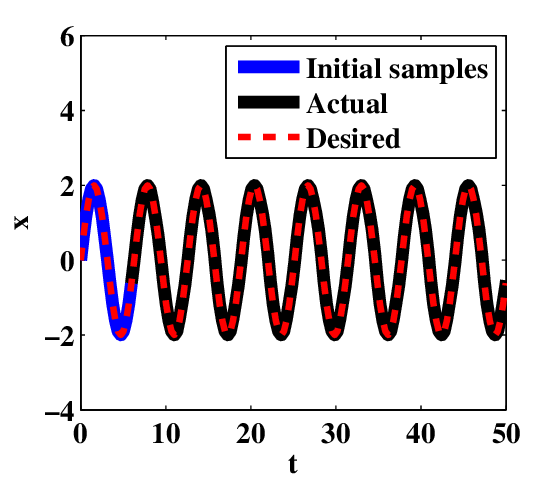}} & \subfigure[][]{\includegraphics[width = 1.5in]{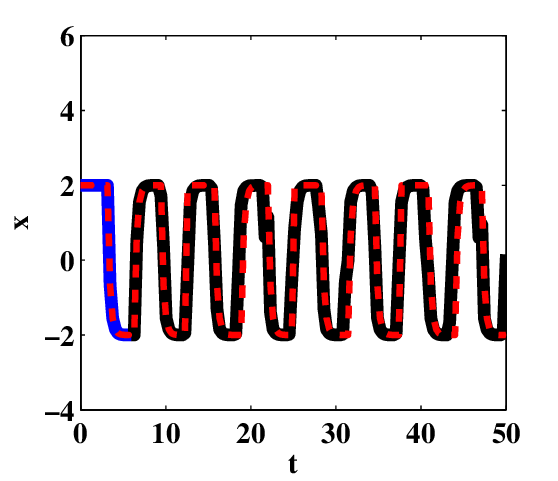}}  \\
    \subfigure[][]{\includegraphics[width = 1.5in]{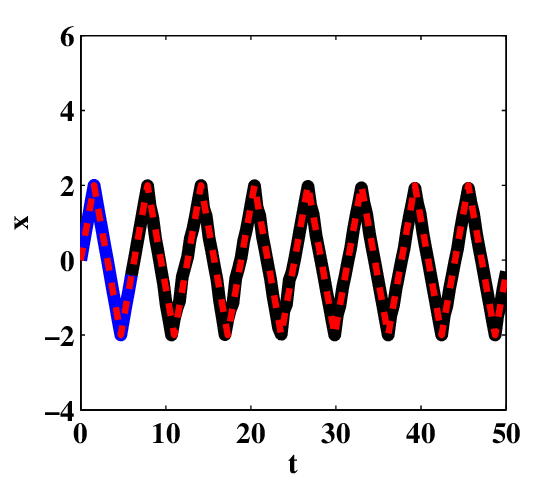}} & \subfigure[][]{\includegraphics[width = 1.5in]{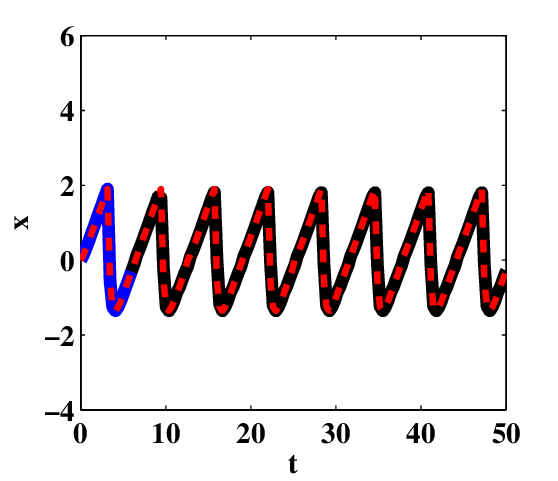}}
\end{tabular}
\caption{Learning four different patterns. \textbf{a} Sine wave. \textbf{b} Square wave. \textbf{c} Triangle wave. \textbf{d} Sawtooth wave.}
\label{fig_5}
\end{figure}

Fig. \ref{fig_5} shows the generated periodic patterns, after receiving the first period. It is shown that the network generates three periods of each pattern, after receiving the first period.

\begin{figure}
\centering
\begin{tabular}{cc}
    \subfigure[][]{\includegraphics[width = 1.5in]{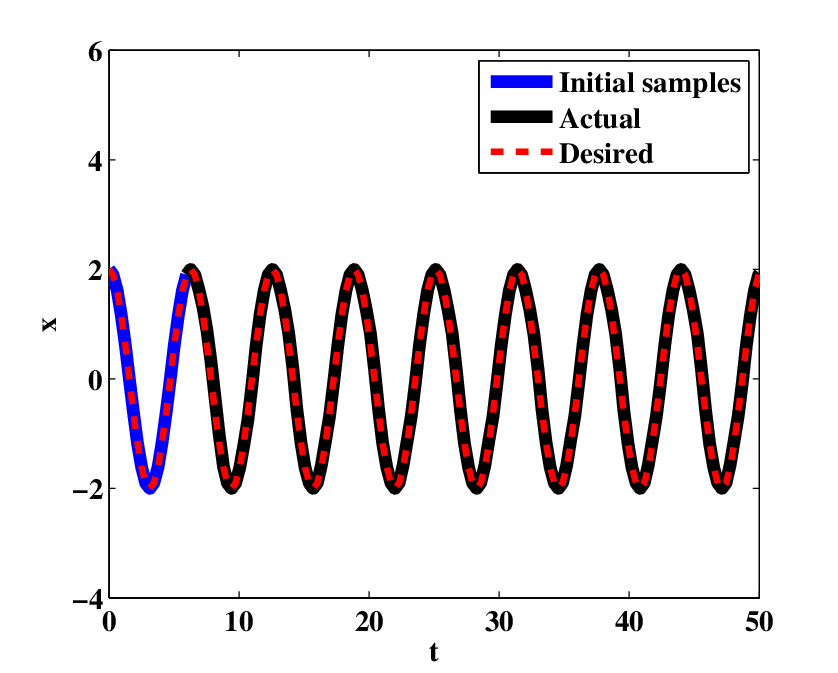}} & \subfigure[]{\includegraphics[width = 1.5in]{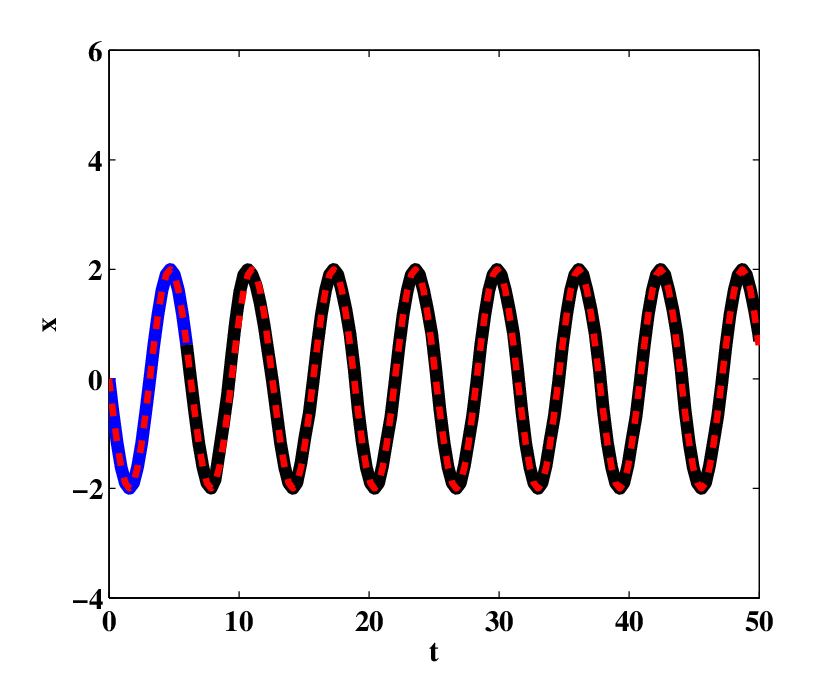}}  \\
\end{tabular}
\caption{Generating learned periodic signal with phase shift. \textbf{a} $\phi = \pi/2$. \textbf{b} $\phi = \pi$.}
\label{fig_6}
\end{figure}

Since the learned sequences are periodic, the output of the \textit{sequence identifier} part is shift invariant. Indeed, it is expected that the network would be able to produce learned patterns with phase shift. Fig. \ref{fig_6} shows the output of the network for sine wave with different phase shifts ($\phi$) including $\pi/2$ (cosine), $\pi$ (inverted sine). It is shown that the proposed network is able to reproduce them with a negligible error.

\begin{figure}[!t]
\centering

\begin{tabular}{ccc}
    \subfigure[][]{\includegraphics[width = 1.3in]{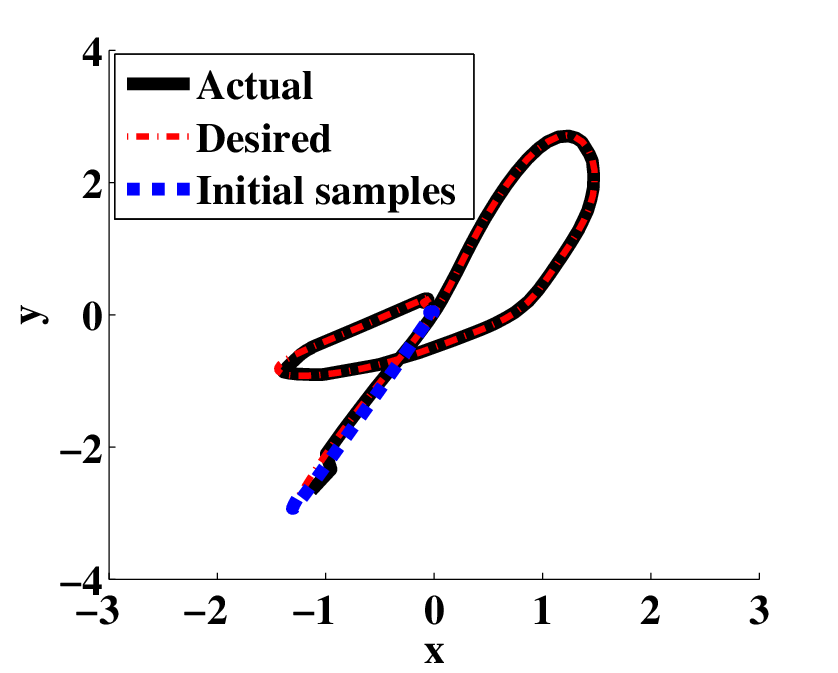}} & \subfigure[][]{\includegraphics[width = 1.3in]{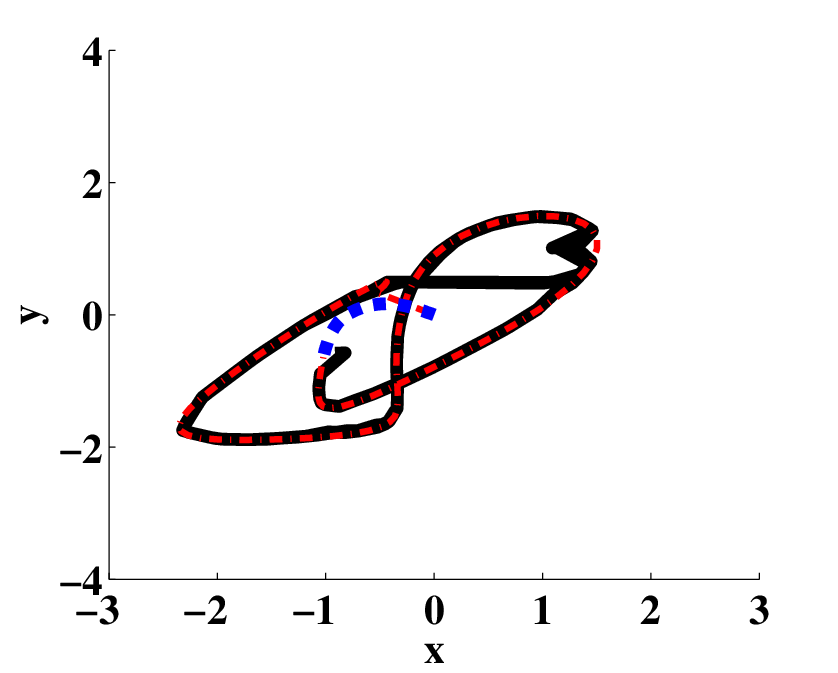}}& \subfigure[][]{\includegraphics[width = 1.3in]{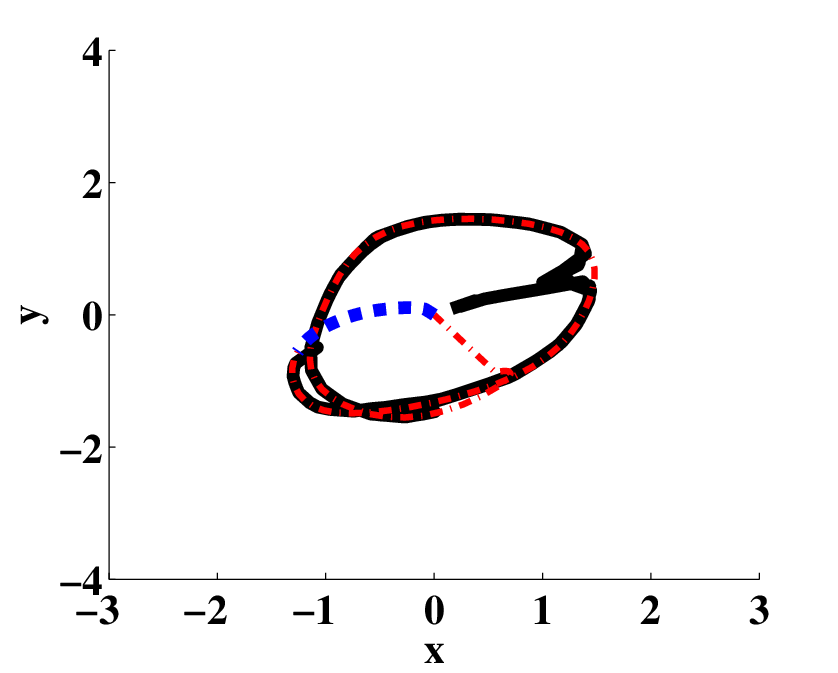}} \\
        \subfigure[][]{\includegraphics[width = 1.3in]{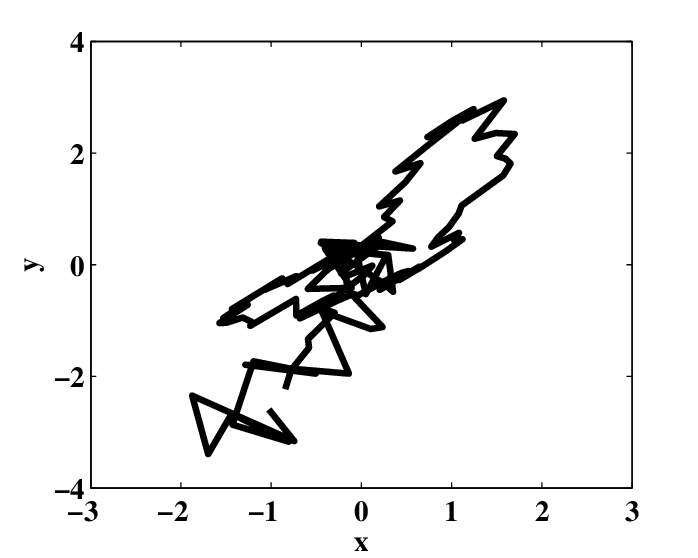}} & \subfigure[][]{\includegraphics[width = 1.3in]{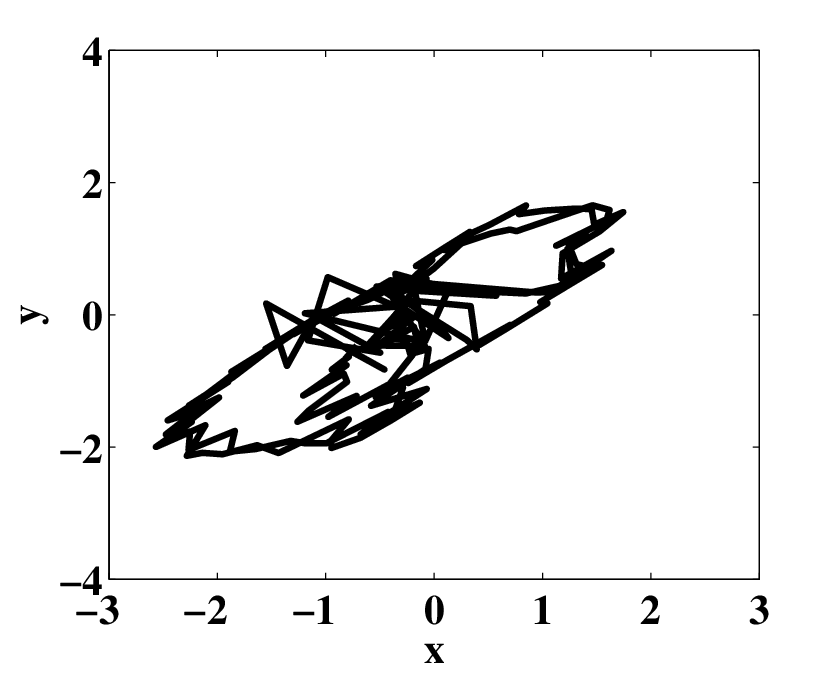}}& \subfigure[][]{\includegraphics[width = 1.3in]{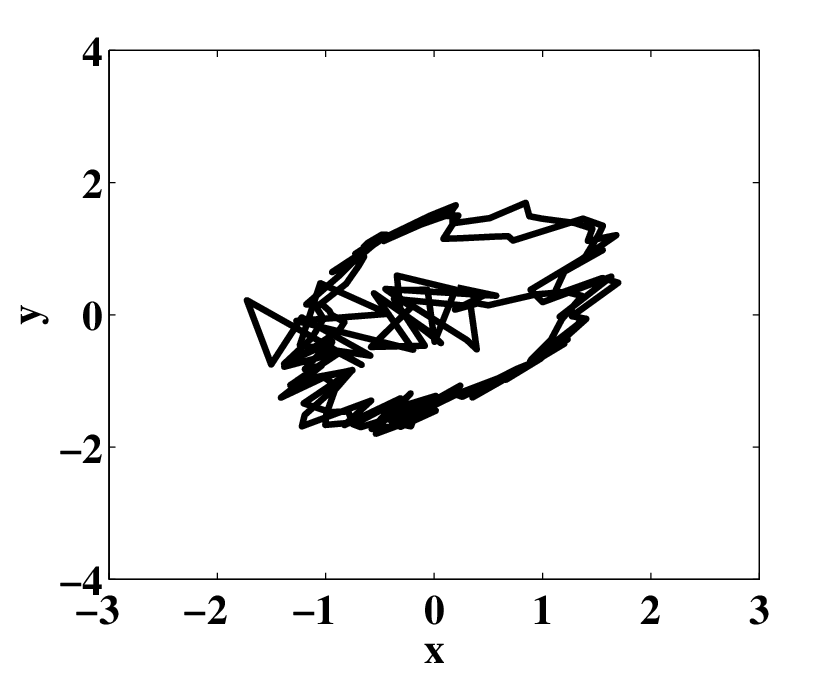}}\\
        \subfigure[][]{\includegraphics[width = 1.3in]{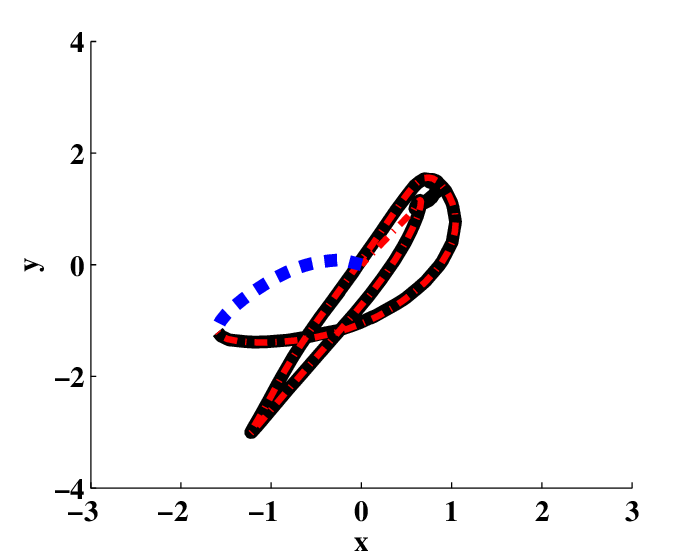}} & \subfigure[][]{\includegraphics[width = 1.3in]{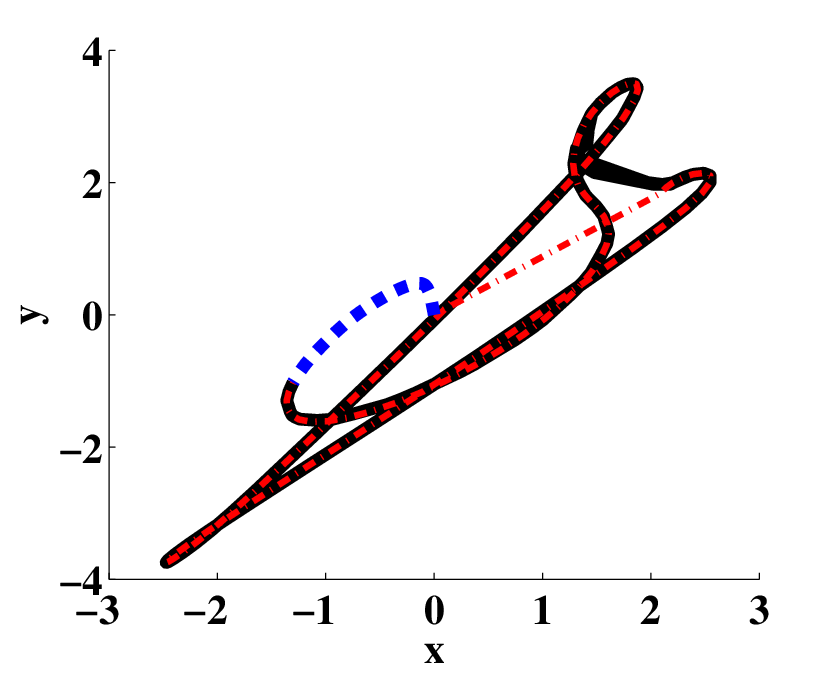}}& \subfigure[][]{\includegraphics[width = 1.3in]{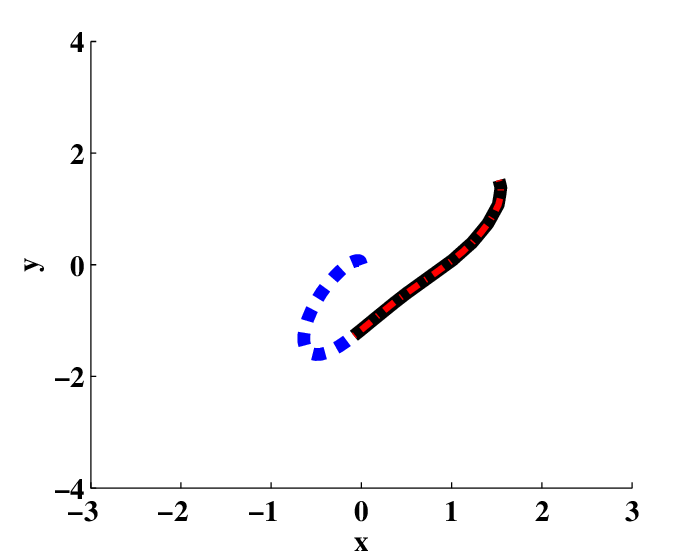}} \\
        \subfigure[][]{\includegraphics[width = 1.3in]{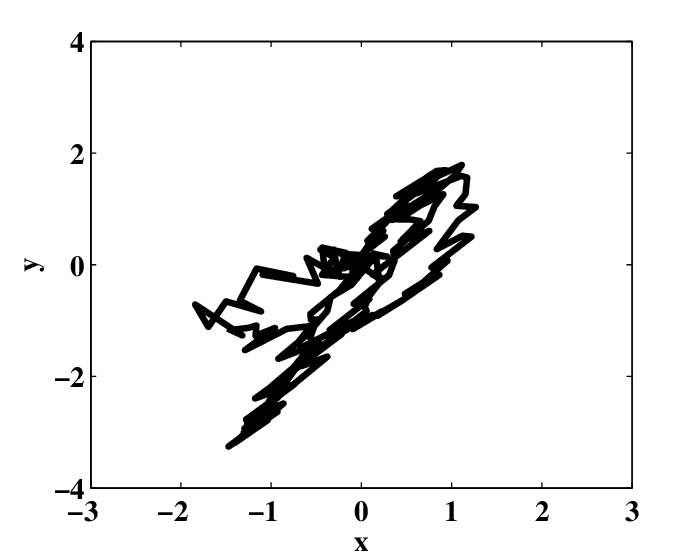}} & \subfigure[][]{\includegraphics[width = 1.3in]{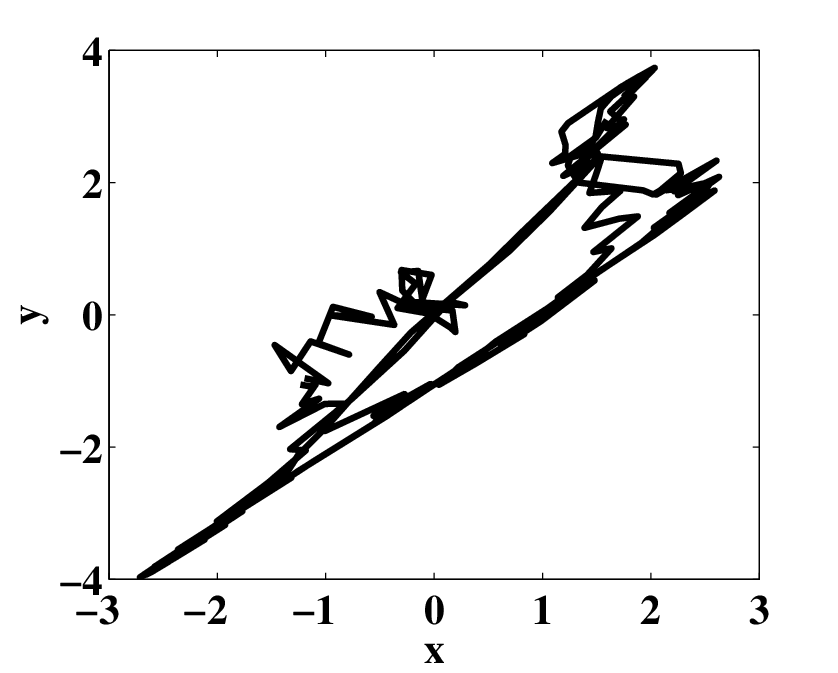}}& \subfigure[][]{\includegraphics[width = 1.3in]{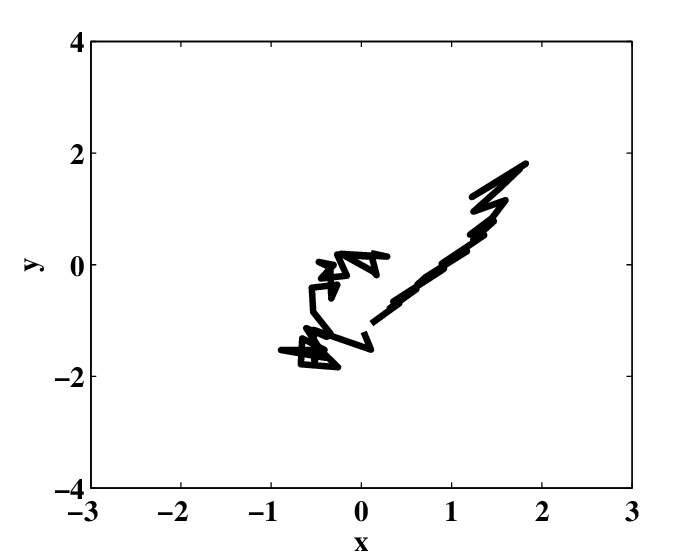}}\\
        \subfigure[][]{\includegraphics[width = 1.3in]{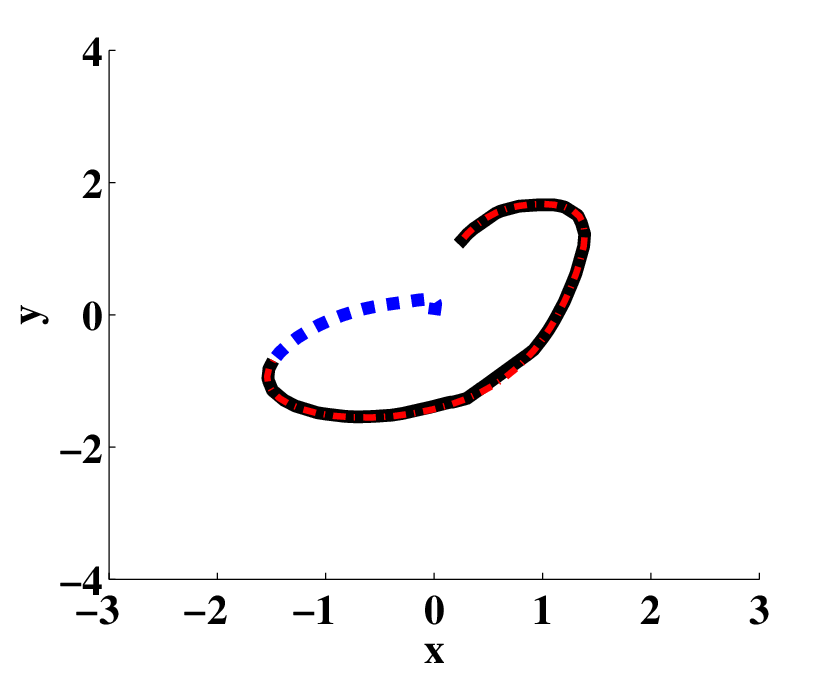}} & \subfigure[][]{\includegraphics[width = 1.3in]{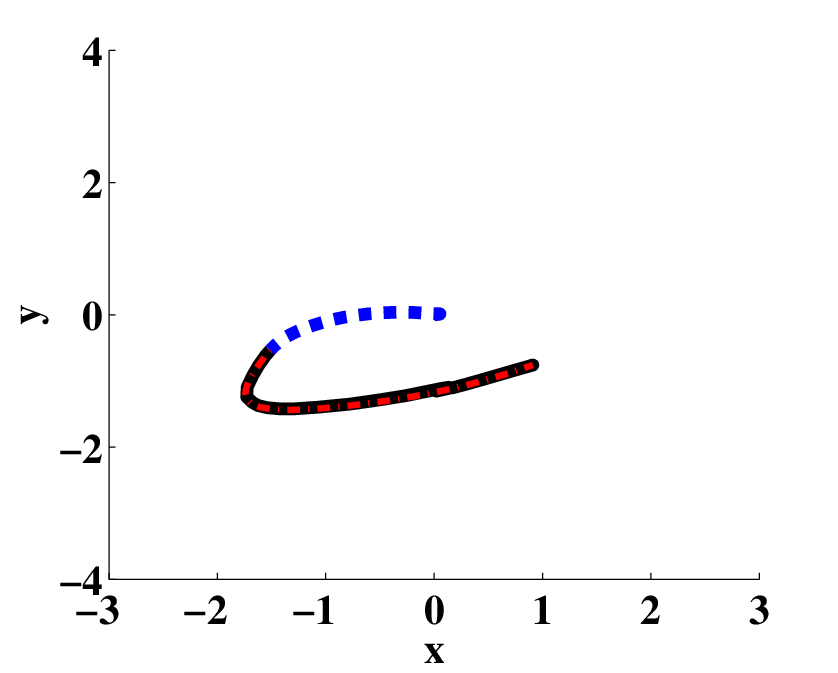}}& \subfigure[][]{\includegraphics[width = 1.3in]{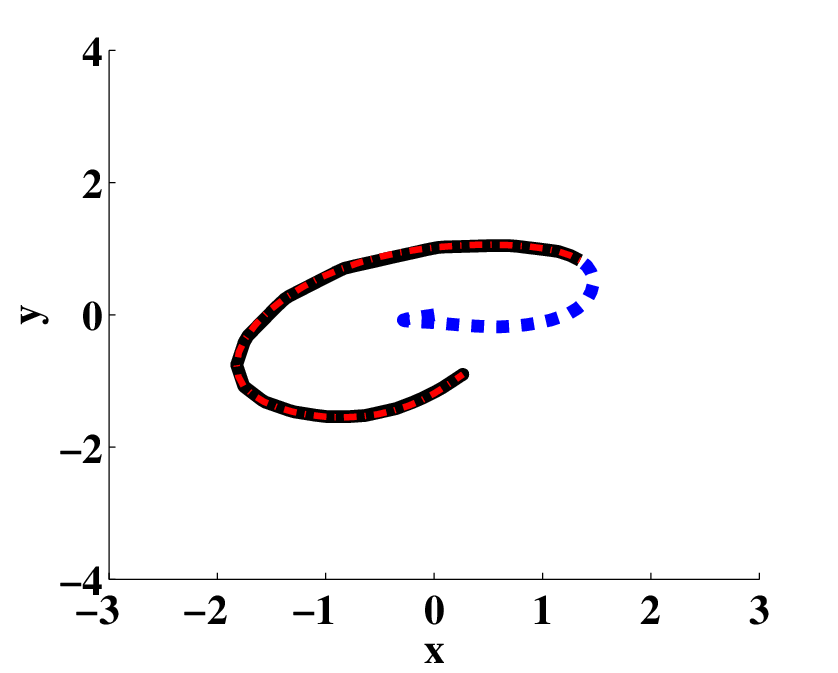}}\\
        \subfigure[][]{\includegraphics[width = 1.3in]{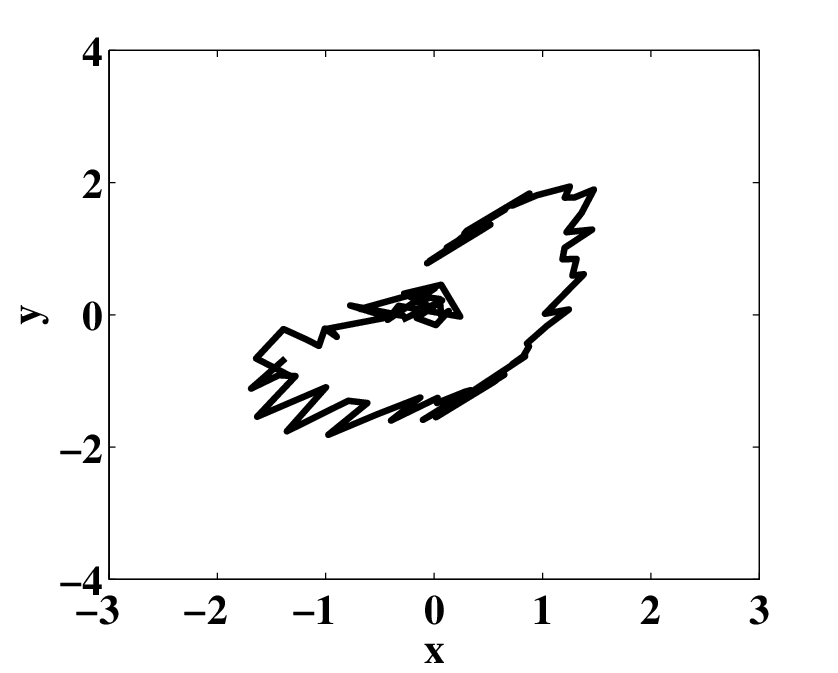}} & \subfigure[][]{\includegraphics[width = 1.3in]{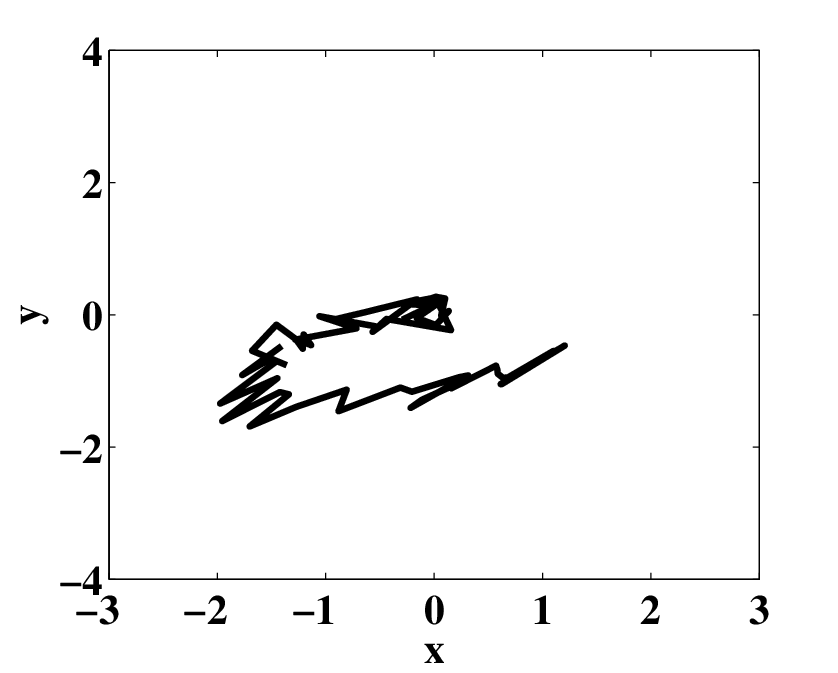}}& \subfigure[][]{\includegraphics[width = 1.3in]{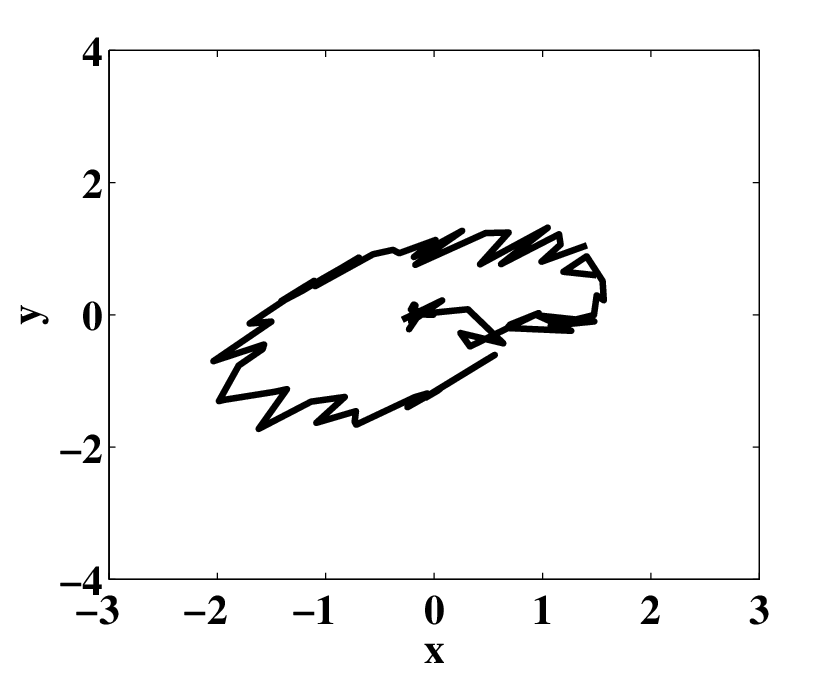}}

\end{tabular}

\caption{Learning and generating different trajectories by the network. The first, third and fifth rows compare the desired and actual sequences. The second, fourth and sixth rows, show the corresponding noisy inputs to the network.}
\label{fig_7}
\end{figure}

\subsection{Different character trajectories}
In this experiment, the proposed network is utilized to learn and generate nine characters including 'a', 'c', 'd', 'e', 'g', 'o', 'p', 'q', and 'u' from UCI character trajectories data set. These selected characters have some similarities; for example 'a', 'c', 'g', 'q', 'u' and 'o' beginning curves are very similar. The network should receive $T = 30$ samples, and generate the remaining 150 samples. Table \ref{table_3} includes different parameters values. To show the ability of the fuzzy network to encounter uncertainty, all input samples are integrated with random noisy samples. After learning, number of fuzzy sets of the \textit{sequence fuzzy sets layer} is 9 and number of fuzzy sets of the \textit{sample fuzzy sets layer} is 172.

\begin{table}[t]
    \caption[c]{Parameters values for learning different character trajectories}
    \begin{center}
    \begin{tabular}{|c|c|c|c|c|c|c|c|}
        \hline
         $iter_{max}$ & $\theta_1$ & $\theta_2$ & $\theta_3$ & $\sigma$ & $T$ & $d$ &$n$\\
         \hline
         20 & 0.2 & 0.2 & 0.01 & 0.2 & 30 & 30 & 2  \\
         \hline
    \end{tabular}
    \end{center}
 \label{table_3}
 \end{table}

Fig. \ref{fig_7} shows the input samples to the network and compares the desired and generated trajectories. All character trajectories are begun from a same point (0,0). Uniform random noise in range (-0.3,0.3) is added to the input samples. Although the input to network is noisy (even rows in Fig. \ref{fig_7}), the network is able to identify and generate each character trajectory with a little error. Therefore, the network was able to learn and generate different sequences.

\section{Conclusions}
\label{section4}
In this paper, a novel self-organizing fuzzy neural network, with a new architecture is proposed for sequence learning. The network composed of two different parts, \textit{sequence identifier}, to recognize different sequences, and \textit{sequence locator} to detect the current location of the sequence. The network is able to identify each learned sequence by receiving $T$ initial samples. Afterward, it would be able to produce the next sample based on the location of the current sample.

The learning algorithm is a two step gradual learning method. Firstly, the fuzzy sets are initialized by correct sequence samples. In this step, new fuzzy sets are added based on the amount of fuzzy sets coverage. If the current set of fuzzy sets does not cover the current sample properly, a new fuzzy set is added. After initializing fuzzy sets, the parameters are fine-tuned by a gradient descent based method, while feeding back the current output of the network as the next input.

The proposed network is used in different sequence learning problems. First, it is utilized to learn two different sequences with a similar part. The network can learn these trajectories successfully because, it is designed to identify each sequence based on initial samples. It is also utilized as pattern generator, to learn and generate different patterns periodically. Finally, as a real world problem, different hand writing character trajectories are learned and reproduced by the proposed network successfully.

The proposed network is an FNN with uncorrelated fuzzy sets. Considering correlations among different dimensions in defining fuzzy sets could be useful to decrease number of required fuzzy sets \citep{Ebadzadeh15,Ebadzadeh2017}. Extending the proposed network with correlated fuzzy rules is proposed as a future work of the current study.

\bibliographystyle{plainnat}
\bibliography{ref}

\end{document}